\documentclass[11pt,a4paper]{article}
\usepackage{acl2021}
\usepackage{times}
\usepackage{latexsym}

\usepackage{xurl}

\usepackage{graphicx}
\graphicspath{ {./images/} }

\usepackage{microtype}
\aclfinalcopy
\usepackage{xr}
\makeatletter
\newcommand*{\addFileDependency}[1]{
  \typeout{(#1)}
  \@addtofilelist{#1}
  \IfFileExists{#1}{}{\typeout{No file #1.}}
}
\makeatother
\usepackage{xcolor}
\usepackage{colortbl}

\newcommand*{\myexternaldocument}[1]{
    \externaldocument{#1}
    \addFileDependency{#1.tex}
    \addFileDependency{#1.aux}
}

\myexternaldocument{hpo4hf_appendix}
\usepackage{multirow}

\usepackage{amsmath}
\DeclareMathOperator*{\argmax}{arg\,max}

\title{An Empirical Study on Hyperparameter Optimization for Fine-Tuning Pre-trained Language Models}

\author{Xueqing Liu \\
  Stevens Institute of Technology \\
  \texttt{xueqing.liu@stevens.edu} \\\And
  Chi Wang \\
  Microsoft Research\\
  \texttt{wang.chi@microsoft.com} \\}

\date{}
\begin{document}
\maketitle
\begin{abstract}


The performance of fine-tuning pre-trained language models largely depends on the hyperparameter configuration. 
In this paper, we investigate the performance of modern hyperparameter optimization methods (HPO) on fine-tuning pre-trained language models. 
First, we study and report three HPO algorithms' performances on fine-tuning two state-of-the-art language models on the GLUE dataset. We find that using the same time budget, HPO often fails to outperform grid search due to two reasons: insufficient time budget and overfitting. We propose two general strategies and an experimental procedure to systematically troubleshoot HPO's failure cases. By applying the procedure, we observe that HPO can succeed with more appropriate settings in the search space and time budget; however, in certain cases overfitting remains. Finally, we make suggestions for future work. Our implementation can be found in \url{https://github.com/microsoft/FLAML/tree/main/flaml/nlp/}.

\end{abstract}

\section{Introduction}
\label{sec:intro}

In the recent years, deep learning and pre-trained language models~\cite{devlin2018bert,liu2019roberta,clark2020electra,he2020deberta} have achieved great success in the NLP community. It has now become a common practice for researchers and practitioners to fine-tune pre-trained language models in down-stream NLP tasks. For example, the HuggingFace transformers library~\cite{wolf2019huggingface} was ranked No.1 among the most starred NLP libraries on GitHub using Python\footnote{https://github.com/EvanLi/Github-Ranking/blob/master/Top100/Python.md}. 

Same as other deep learning models, the performance of fine-tuning pre-trained language models largely depends on the hyperparameter configuration. A different setting in the hyperparameters may cause a significant drop in the performance, turning a state-of-the-art model into a poor model. Methods for tuning hyperparameters can be categorized as (1) traditional approaches such as manual tuning and grid search, and (2) automated HPO methods such as random search and Bayesian optimization (BO). Manual tuning often requires a large amount of manual efforts; whereas grid search often suffers from lower efficiency due to the exponential increase in time cost with the number of hyperparameters. Automated HPO methods were proposed to overcome these disadvantages. 
Recently, automated HPO methods also become increasingly popular in the NLP community~\cite{zhang2020reproducible,dodge2019show}. For example, Bayesian optimization (BO)~\cite{zhang2020reproducible} and Population-based Training~\cite{jaderberg2017population} both prove to be helpful for improving the performance of the transformer model~\cite{vaswani2017attention} for neural machine translation. The HuggingFace library has also added native supports for HPO in a recent update (version 3.1.0, Aug 2020).

With improved supports, users can now easily access a variety of HPO methods and apply them to their fine-tuning tasks. However, the effectiveness of this step is less understood. To bridge this gap, in this paper, we propose an experimental study for fine-tuning pre-trained language models using the HuggingFace library. This study is motivated by the following research questions: First, can automated HPO methods outperform traditional tuning method such as grid search? Second, on which NLP tasks do HPO methods work better? Third, if HPO does not work well, how to troubleshoot the problem and improve its performance? 

To answer these questions, we start from a simple initial study (Section~\ref{sec:init_electra_exp}) by examining the performance of three HPO methods on two state-of-the-art language models on the GLUE dataset. The time budget for HPO in the initial study is set to be the same as grid search. Results of the initial study show that HPO often fails to match grid search's performance. The reasons for HPO's failures are two folds: first, the same budget as grid search may be too small for HPO; second, HPO overfits the task. With these observations, we propose two general strategies for troubleshooting the failure cases in HPO as well as an overall experimental procedure (Figure~\ref{fig:procedure}). By applying the procedure (Section~\ref{sec:test_procedure}), we find that by controlling overfitting with reduced search space and using a larger time budget, HPO has outperformed grid search in more cases. However, the overfitting problem still exists in certain tasks even when we only search for the learning rate and batch size. Finally, we make suggestions for future work (Section~\ref{sec:discussion}).

The main contributions of this work are:

\begin{itemize}
    \item We empirically study the performance of three HPO methods on two pre-trained language models and on the GLUE benchmark;
    \item We design an experimental procedure which proves useful to systematically troubleshoot the failures in HPO for fine-tuning;
    \item We report and analyze the execution results of the experimental procedure, which sheds light on future work;
\end{itemize}



\section{Definition of HPO on Language Model Fine-Tuning}
\label{sec:bg}

\noindent Given a pre-trained language model, a fine-tuning task, and a dataset containing \(D_{train}, D_{val}, D_{test}\), the goal of a hyperparameter optimization algorithm is to find a hyperparameter configuration $\mathbf{c}$, so that when being trained under configuration $\mathbf{c}$, the model's performance on a validation set $D_{val}$ is optimized. Formally, the goal of HPO is to find
\begin{equation*}
\mathbf{c}^* = \argmax_{\mathbf{c}\in \mathcal{S}} f(\mathbf{c}, D_{train}, D_{val}) 
\end{equation*}
where $\mathcal{S}$ is called the \emph{search space} of the HPO algorithm, i.e., the domain where the hyperparameter values can be chosen from. The function $f(\cdot, \cdot, \cdot)$ is called the evaluation protocol of HPO, which is defined by the specific downstream task. For example, many tasks in GLUE define $f$ as the validation accuracy. If a task has multiple protocols, we fix $f$ as one of them\footnote{There are 3 GLUE tasks with multiple validation scores: MRPC, STS-B, and QQP (not studied). For MRPC we optimize the validation accuracy, and for STS-B we optimize the Pearson score on the validation set. }. After finding $\mathbf{c}^*$, the performance of HPO will be evaluated using the  performance of the model trained with $\mathbf{c}^*$ on the \emph{test} set $D_{test}$. 

To fairly compare the performances of different HPO algorithms, the above optimization problem is defined with a constraint in the maximum running time of the HPO algorithm, which we call the \emph{time budget} for the algorithm, denoted as $B$. Under budget $B$, the HPO algorithm can try a number of configurations $\mathbf{c}_1, \mathbf{c}_2, \cdots, \mathbf{c}_n$. The process of fine-tuning with configuration $\mathbf{c}_i$ is called a \emph{trial}. Finally, we call the process of running an HPO algorithm $A$ once \emph{one HPO run}. 


\section{Factors of the Study}
\label{sec:env_factors}

\begin{table*}[!h]
\begin{center}
\begin{tabular}{p{3cm}p{3cm}p{3cm}p{2.5cm}p{2cm}}
\hline { Hyperparameter } & { Electra-grid } & Electra-HPO & RoBERTa-grid & RoBERTa-HPO \\ \hline
learning rate &\{3e-5,1e-4,1.5e-4\} & log((2.99e-5,1.51e-4)) & \{1e-5,2e-5,3e-5\} & (0.99e-5,3.01e-5)\\
warmup ratio  & 0.1 & (0, 0.2) & 0.06 & (0, 0.12) \\
attention dropout & 0.1 & (0, 0.2) & 0.1 & (0, 0.2) \\
hidden dropout & 0.1 &  (0, 0.2) & 0.1 & (0, 0.2)\\
weight decay & 0 & (0, 0.3) & 0.1 & (0, 0.3)\\
batch size & 32 & \{16, 32, 64\} & \{16, 32\} & \{16, 32, 64\} \\
epochs & 10 for RTE/STS-B, 3 for other & 10 for RTE/STS-B, 3 for other & 10 & 10\\
\hline
\end{tabular}
\end{center}
\caption{The search space for grid search and HPO methods in this paper. For grid search, we adopt the search spaces from the Electra~\cite{clark2020electra} and RoBERTa~\cite{liu2019roberta} paper. For each model, we expand the grid search space to a larger, simple search space for HPO. \label{tab:search_space} }
\end{table*}

In this paper, we conduct an empirical study to answer the research questions in Section~\ref{sec:intro}. First, can automated HPO methods outperform grid search? The answer to this question depends on multiple factors, i.e., the NLP task on which HPO and grid search are evaluated, the pre-trained language model for fine tuning, the time budget, the search space for grid search and HPO algorithm, and the choice of HPO algorithm. To provide a comprehensive answer, we need to enumerate multiple settings for these factors. However, it is infeasible to enumerate all possible settings for each factor. For instance, there exist unlimited choices for the search space. To accomplish our research within reasonable computational resources\footnotemark, for each factor, we only explore the most straight-foward settings. For example, the search space for grid search is set as the default grid configuration recommended for fine-tuning (Table~\ref{tab:search_space}), and the search space for HPO is set as a straightforward relaxation of the grid configuration. We explain the settings for each factor in details below. \\

\footnotetext{Our experiments were run on two GPU servers, server 1 is equipped with 4xV100 GPUs (32GB), and server 2 is a DGX server equipped with 8xV100 GPUs (16GB). To avoid incomparable comparisons, all experiments on QNLI and MNLI are run exclusively on server 2, and all other experiments are run exclusively on server 1. To speed up the training, we use fp16 in all our experiments. To guarantee the comparability between different HPO methods, all trials are allocated exactly 1 GPU and 1 CPU. As a result, all trials are executed in the single-GPU mode and there never exist two trials sharing the same GPU. }

\noindent \textbf{NLP Tasks}. To study HPO's performance on multiple NLP tasks, we use the 9 tasks from the GLUE (General Language Understanding Evaluation) benchmark~\cite{wang2018glue}. \\

\noindent \textbf{Time Budget}. We focus on a low-resource scenario in this paper. To compare the performance of grid search vs. HPO, we first allocate the same time budget to HPO as grid search in our initial comparative study (Section~\ref{sec:init_electra_exp}). If HPO does not outperform grid search, we increase the time budget for HPO. We require that each HPO run takes no more than 8 GPU hours with the NVIDIA Tesla V100 GPU under our setting. We prune a task if the time for grid search exceeds two hours. A complete list of the time used for each remaining task can be found in Table~\ref{tab:time_resource}. \\

\begin{table}[h]
\begin{center}
\begin{tabular}{lp{0.7cm}p{1cm}|p{1.2cm}p{1cm}}
\hline { NLP task } & { Electra } & epoch & RoBERTa & epoch \\ \hline
WNLI & 420 & 3 & 660 & 10 \\
RTE & 1000 & 10 &  720 & 10\\
MRPC & 420 & 3& 720 & 10\\
CoLA  & 420 & 3& 1200 & 10 \\
STS-B & 1200 & 10& 1000  & 10\\
SST & 1200& 3 &  \textbf{7800} & -\\
QNLI & 1800 & 3& - \\
QQP & \textbf{7800} & 3& -& -\\
MNLI & 6600 & 3& -& - \\
\hline
\end{tabular}
\end{center}
\caption{The running time of grid search for each task (in seconds) and the corresponding number of epochs. \label{tab:time_resource} }
\end{table}

\noindent \textbf{Pre-trained Language Models}. In this paper, we focus on two pre-trained language models: the Electra-base model~\cite{clark2020electra}, and the RoBERTa-base model~\cite{liu2019roberta}. Electra and RoBERTa are among the best-performing models on the leaderboard of GLUE as of Jan 2021\footnote{\url{www.gluebenchmark.com}}. Another reason for choosing the two models is that they both provide a simple search space for grid search, and we find it helpful to design our HPO search space on top of them. We use both models' implementations from the transformers library~\cite{wolf2019huggingface} (version = 3.4.0). Among all the different sizes of RoBERTa and Electra (large, base, small), we choose the base size, because large models do not fit into our 2-hour budget\footnote{Our empirical observation shows that the large models take 1.5 to 2 times the running time of the base models. }. With the 2-hour time constraint, we prune tasks where grid search takes longer than two hours. For Electra, QQP is pruned, whereas for RoBERTa, SST, QNLI, QQP, MNLI are pruned. \\

\noindent \textbf{Search Space for Grid Search and HPO}. It is generally difficult to design an HPO search space from scratch. In our problem, this difficulty is further amplified with the limited computational resources. Fortunately, most papers on pre-trained language models recommend one or a few hyperparameter configurations for fine-tuning. We use them as the configurations for grid search. For HPO, the performance depends on the search space choice, e.g., it takes more resources to explore a large space than a smaller space close to the best configuration. Due to the time budget limits, we focus on a small space surrounding the recommended grid search space,  as shown in Table~\ref{tab:search_space}.\footnote{The grid search spaces in Table~\ref{tab:search_space} are from Table 7 of Electra and Table 10 of RoBERTa. For Electra, we fix the hyperparameters for Adam; we skip the layer-wise learning rate decay because it is not supported by the HuggingFace library. While Electra's original search space for learning rate is [3e-5, 5e-5, 1e-4, 1.5e-4], we have skipped the learning rate 5e-5 in our experiment. } More specifically, we convert the \emph{learning rate}, \emph{warmup ratio}, \emph{attention dropout}, and \emph{hidden dropout} to a continuous space by expanding the grid space. For \emph{weight decay}, since the recommended configuration is 0, we follow Ray Tune's search space and set the HPO space to (0, 0.3)~\cite{ray_blog}. For \emph{epoch number}, most existing work uses an integer value between 3 and 10~\cite{clark2020electra,liu2019roberta,dai2020funnel}, resulting in a large range of space we can possibly search. To reduce the exploration required for HPO, we skip expanding the search space for epoch number and fix it to the grid configuration. \\

\noindent \textbf{HPO Algorithms}. We compare the performance between grid search and three HPO algorithms: random search~\cite{bergstra2012random}, asynchronous successive halving (ASHA)~\cite{Li2020}, and Bayesian Optimization~\cite{akiba2019optuna}+ASHA. We use all HPO methods' implementations from the Ray Tune library~\cite{liaw2018tune} (version 1.2.0). We use BO (with TPE sampler) together with the ASHA pruner, because with the small time budget, BO without the pruner reduces to random search. As fine-tuning in NLP usually outputs the checkpoint with the \emph{best} validation accuracy, we also let the HPO methods output the \emph{best} checkpoint of the best trial. This choice is explained in more details in Appendix~\ref{file2:sec:checkpoint}. \\

\begin{table*}[!h]
\begin{center}
\begin{tabular}{lcccccccc}
\hline   &  \textbf{WNLI} &  \textbf{RTE}  &   \textbf{MRPC}  &  \textbf{CoLA}  & \textbf{STS-B} &  \textbf{SST}  &  \textbf{QNLI} &  \textbf{MNLI}  \\
\hline  
\multicolumn{8}{l}{\emph{Electra-base, validation }} \\
grid  & 56.3 & {\bf 84.1} & 92.3/89.2 & 67.2 & {\bf 91.5/91.4} & {\bf 95.1} & {\bf 93.5} &  88.6  \\
RS & 56.8 & 82.2 & {\bf 93.0/90.4} & 68.8 & 90.1/90.2 & 94.7 & 93.0 &  88.9\\ 
RS+ASHA & 57.2 & 80.3 & 93.0/90.3 & 67.9 & 91.4/91.3 & 94.9 & 93.1 &  88.6\\
BO+ASHA & {\bf 58.2} & 82.6 & {\bf 93.1/90.4} & {\bf 69.4} & 91.5/91.3 & 94.7 & 93.1 & {\bf 89.2}\\
\hline
\multicolumn{8}{l}{\it Electra-base, test} & \\
grid  & {\bf 65.1} & {\bf 76.8} & {\bf 91.1/87.9} & 58.5 & {\bf 89.7/89.2} & {\bf 95.7} & {\bf 93.5} & 88.3\\
RS & 64.4 & 75.6 & 90.7/87.5 & 63.0 & 88.0/87.6 & 95.1 & 93.0 & {\bf 88.7}\\ 
RS+ASHA & 62.6 & 74.1 & 90.6/87.3 & 61.2 & 89.5/89.1 & 94.9 & 92.9 &  88.5\\
BO+ASHA  & 61.6 & 75.1 & 90.7/87.4 & {\bf 64.1} & 89.7/89.1 & 94.8 & 93.0 & {\bf 88.7}\\
\hline
\end{tabular}
\end{center}
\begin{center}
\begin{tabular}{lccccc}
\hline   &  \textbf{WNLI} &  \textbf{RTE}  &   \textbf{MRPC}  &  \textbf{CoLA}  & \textbf{STS-B} \\
\hline  
\multicolumn{5}{l}{\emph{RoBERTa-base, validation }} & \\
grid  & 56.3 & 79.8 & 93.1/90.4 & {\bf 65.1} & 91.2/90.8 \\
RS & {\bf 57.8} & 80.4 & 93.3/90.7 & 64.1 & 91.2/90.9\\ 
RS+ASHA & 57.3 & {\bf 80.8} & 93.4/90.8 & 64.5 & 91.2/90.9\\
BO+ASHA & 56.3 & 80.3 & {\bf 93.7/91.4} & 64.5 & {\bf 91.3/91.0}\\
\hline
\multicolumn{5}{l}{\it RoBERTa-base, test} & \\
grid  & {\bf 65.1} & 73.9 & 90.5/87.1 & {\bf 61.7} & 89.3/88.4\\
RS & 64.9 & 73.5 & 90.1/86.7 & 59.1 & {\bf 89.3/88.6}\\ 
RS+ASHA & {\bf 65.1} & {\bf 74.1} & {\bf 90.6/87.3} & 59.4 & 89.1/88.3\\
BO+ASHA  & {\bf 65.1} & 73.3 & 90.4/87.2 & 60.1 & 89.1/88.4\\
\hline
\end{tabular}
\end{center}

\caption{Results of the initial comparative study on Electra (top) and RoBERTa (bottom) by varying the GLUE task and HPO method while fixing the search space and time budget. For each (HPO method, task), we rerun the experiment 3 times and report the average. \label{tab:init_result_electra}}
\end{table*}

\section{Experiment \#1: Comparative Study using 1GST}
\label{sec:init_electra_exp}

As the performance of HPO depends on the time budget, to compare between grid search and HPO, we first conduct an initial study by setting the time budget of HPO to the same as grid search. For the rest of this paper, we use $a$GST to denote that the time budget=$a\times$the running time for grid search.  Table~\ref{tab:init_result_electra} shows the experimental results on Electra and RoBERTa using 1GST. For each (HPO method, NLP task) pair, we repeat the randomized experiments 3 times and report the average scores. We analyze the results in Section~\ref{sec:electra_analysis}. 

\subsection{Analysis of the Initial Results}
\label{sec:electra_analysis}


\noindent\textbf{Electra}. By comparing the performance of grid search and HPO in Table~\ref{tab:init_result_electra} we can make the following findings. First, HPO fails to match grid search's validation accuracy in the following tasks: RTE, STS-B, SST and QNLI. In certain tasks such as QNLI and RTE, grid search outperforms HPO by a large margin. Considering the fact that grid search space is a subspace of the HPO space, this result shows that with the same time budget as grid search (i.e., approximately 3 to 4 trials), it is difficult to find a configuration which works better than the recommended configurations. Indeed, with 3 to 4 trials, it is difficult to explore the search space. Although ASHA and BO+ASHA both search for more trials by leveraging early stopping~\cite{Li2020}, the trial numbers are still limited (the average trial numbers for experiments in Table~\ref{tab:init_result_electra} can be found in Table~\ref{file2:tab:trial_number} of the appendix). Second, among the tasks where HPO outperforms grid search's validation accuracy, there are 2 tasks (WNLI, MRPC) where the test accuracy of HPO is lower than grid search. As a result, the HPO algorithm overfits the validation dataset.
Overfitting in HPO generally happens when the accuracy is optimized on a limited number of validation data points and cannot generalize to unseen test data~\cite{feurer2019hyperparameter}. \cite{zhang2021revisiting} also found that fine-tuning pre-trained language models is prone to overfitting when the number of trials is large, though they do not compare HPO and grid search. Finally, by searching for more trials, ASHA and BO+ASHA slightly outperform random search in the validation accuracy, but their test accuracy is often outperformed by random search. 


\noindent \textbf{RoBERTa}. By observing RoBERTa's results from Table~\ref{tab:init_result_electra}, we can see that the average validation accuracy of HPO outperforms grid search in all tasks except for CoLA. It may look like HPO is more effective; however, most of the individual runs in Table~\ref{tab:init_result_electra} overfit. As a result, HPO for fine-tuning RoBERTa is also prone to overfitting compared with grid search. The complete lists of the overfitting cases in Table~\ref{tab:init_result_electra} can be found in Table~\ref{file2:tab:reduce_sp_analysis_electra} and Table~\ref{file2:tab:reduce_sp_analysis_roberta} of Appendix~\ref{file2:sec:reduce_sp_analysis}. 

\subsection{A General Experimental Procedure for Troubleshooting HPO Failures}
\label{sec:procedure}

\noindent Since Table~\ref{tab:init_result_electra} shows HPO cannot outperform grid search using 1GST, and is prone to overfitting, we propose two general strategies to improve HPO's performance. First, we increase the time budget for HPO so that HPO can exploit the space with more trials. Second, to control overfitting, we propose to reduce the search space. More specifically, we propose to fix the values of certain hyperparameters to the default values in the grid configuration (Table~\ref{tab:init_result_electra}). The reason is that overfitting can be related to certain hyperparameter settings of the model. For example, it was shown in ULMFit~\cite{howard2018universal} that using a non-zero warmup step number can help reduce overfitting. Intuitively, a larger search space is more prone to overfitting. For example, by using a warmup search space = (0, 0.2), the warmup steps in the best trial found by HPO may be much smaller or larger than the steps used by grid search. Other hyperparameters which are related to overfitting of fine-tuning include the learning rate~\cite{smith2017bayesian}, batch size~\cite{smith2017don}, and the dropout rates~\cite{srivastava2014dropout,loshchilov2017decoupled,loshchilov2018fixing}. 

Our proposed procedure for troubleshooting HPO failures is depicted in Figure~\ref{fig:procedure}. Starting from the full search space and 1GST, we test the HPO algorithm for a few times. If any overfitting is observed, we reduce the search space and go back to testing the HPO algorithm again. On the other hand, if no overfitting is observed and HPO also does not outperform grid search, we increase the time budget and also go back to testing the HPO algorithm again. We continue this procedure until any of the following conditions is met: first, HPO successfully outperforms grid search; second, the search space cannot be further reduced, thus HPO overfits the task; third, the time budget cannot be further increased under a user-specified threshold, thus whether HPO can outperform grid search is to be determined for this specific task.


\section{Experiment \#2: Troubleshooting HPO}
\label{sec:test_procedure}

\begin{figure}[h]
\caption{A general experimental procedure for troubleshooting HPO failure cases. \label{fig:procedure} }
\includegraphics[width=\columnwidth]{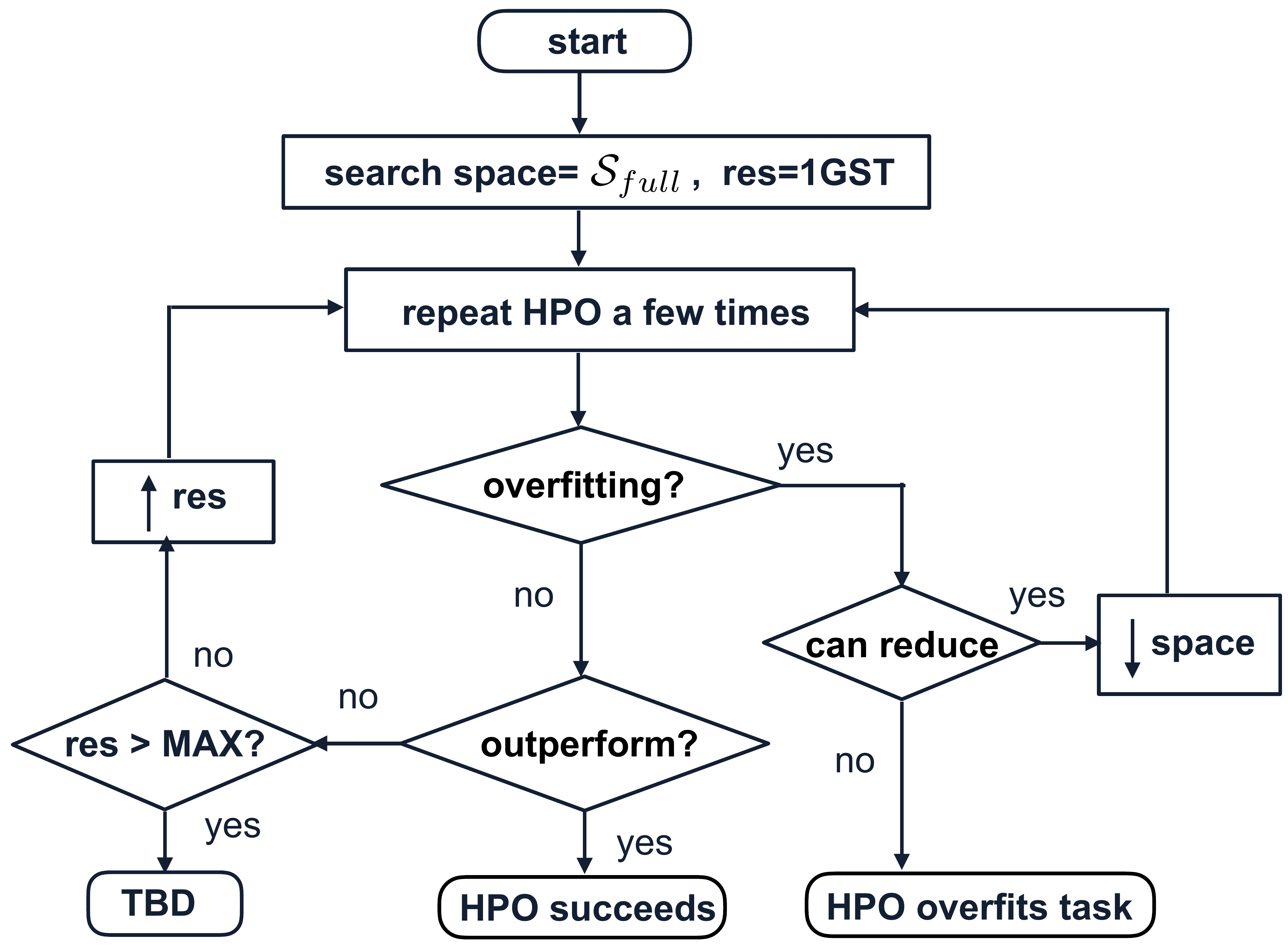}
\end{figure}

In this section, we evaluate the effectiveness of our proposed procedure in Figure~\ref{fig:procedure}. To apply the procedure, we need to further consolidate two components: first, what time budget should we use; second, \emph{which hyperparameter} to fix for reducing the search space. For the first component, we use a relatively small list for time budget options \{1GST, 4GST\}. For the second component, it is difficult to guarantee to reduce overfitting by fixing a specific hyperparameter to its grid search values. When choosing the hyperparameter to fix, we refer to the configurations of the best trials which cause the HPO results to overfit. 

\subsection{Choosing the Hyperparameter to Fix}
\label{sec:reduce_sp_electra}

\textbf{Electra}. To decide which hyperparameter to fix, we examine the best trial's configuration for the overfitting HPO runs (compared with the grid search performance). If there is a pattern in a certain hyperparameter of all these configurations (e.g., warmup ratio below 0.1 for Electra), by fixing such hyperparameters to the values of grid search, we can exclude the other values which may be related to overfitting. We apply this analytical strategy to the initial Electra results in Table~\ref{tab:init_result_electra}. Among the 72 runs, 9 runs overfit compared with grid search. For each run, we list the hyperparameter configurations of the best trial in Table~\ref{file2:tab:reduce_sp_analysis_electra} of Appendix~\ref{file2:sec:reduce_sp_analysis}. For Electra, we have skipped showing weight decay in Table~\ref{file2:tab:reduce_sp_analysis_electra}, because the HPO configuration is never smaller than the grid configuration, thus does not affect the result of the analysis. For comparative purpose, we also list the hyperparameter values of the best trial in grid search. To improve the readability of Table~\ref{file2:tab:reduce_sp_analysis_electra}, we use 4 different colors (defined in Appendix~\ref{file2:sec:reduce_sp_analysis}) to denote the comparison between values of the best trial in HPO and values of the best trial in grid search. 

From Table~\ref{file2:tab:reduce_sp_analysis_electra}, we observe that the \emph{warmup ratio}s are often significantly lower than 0.1. We skip the analysis on learning rate because its search space (log((2.99e-5,1.51e-4))) cannot be further reduced without losing coverage of the grid configurations or continuity; we also skip weight decay because any trial's value cannot be smaller than 0. Following this empirical observation, we hypothesize that fixing the warmup ratio to 0.1 can help reduce overfitting in Electra. We use $\mathcal{S}_{full}$ to denote the original search space and $\mathcal{S}_{-wr}$ to denote the search space by fixing the warmup ratio to 0.1. If HPO overfits in both $\mathcal{S}_{full}$ and $\mathcal{S}_{-wr}$, the procedure will reduce the search space to the minimal continuous space $\mathcal{S}_{min}$ containing the grid search space, which searches for the learning rate only. \\

\noindent \textbf{RoBERTa}. We apply the same analytical strategy to the RoBERTa results in Table~\ref{tab:init_result_electra} and show the hyperparameters of the best trials in Table~\ref{file2:tab:reduce_sp_analysis_roberta}. For RoBERTa, we propose to fix the values of two hyperparameters at the same time: the warmup ratio and the hidden dropout. We denote the search space after fixing them as $\mathcal{S}_{-wr-hdo}$. If HPO overfits in both $\mathcal{S}_{full}$ and $\mathcal{S}_{-wr-hdo}$, the procedure will reduce the search space to $S_{min}$ which contains the learning rate and batch size only. 

\subsection{Execution Results of the Procedure}
\label{sec:procedure_result_electra}

In this section, we apply the troubleshooting procedure on the initial HPO results from Table~\ref{tab:init_result_electra} and observe the execution paths. In Table~\ref{file2:tab:electra_procedure_result} and Table~\ref{file2:tab:roberta_procedure_result} of  Appendix~\ref{file2:exec_results_procedure}, we list the full execution results of the procedure for random search and random search + ASHA. Table~\ref{file2:tab:electra_procedure_result}\&\ref{file2:tab:roberta_procedure_result} have included only the tasks where the HPO does not succeed in the initial study. In Table~\ref{file2:tab:electra_procedure_result}\&\ref{file2:tab:roberta_procedure_result}, we show the validation and test accuracy for the three repetitions of HPO runs as well as their average score.\\

\begin{table}[!h]
\begin{center}
\begin{tabular}{l|p{0.30cm}p{0.40cm}|p{0.40cm}p{0.40cm}|p{0.40cm}p{0.40cm}|p{0.40cm}p{0.40cm}}
\hline
&\multicolumn{2}{c|}{round 0} & \multicolumn{2}{c|}{round  1}  &  \multicolumn{2}{c|}{round 2} & \multicolumn{2}{c}{round 3} \\
&val & test & val & test & val & test & val & test \\ \hline
grid&{\bf\textcolor{blue}{84.1}} & {\bf \textcolor{blue}{76.8}} & \multicolumn{2}{c|}{$\uparrow$ res} &  \multicolumn{2}{c|}{$\downarrow$ space} &  \multicolumn{2}{c}{$\downarrow$ space} \\ \hline
rep1&81.9 & 76.1 & \cellcolor{black!60}{84.5} & \cellcolor{black!60}{74.6} & \cellcolor{black!40}{84.1} & \cellcolor{black!40}{76.1} & \cellcolor{black!60}{84.8} &  \cellcolor{black!60}{75.3} \\
rep2&81.6 & 75.1 & 83.8 & 74.5 & 83.0 &  74.0 & 84.1 & 75.7  \\
rep3&83.0 & 75.7 & 83.4 & 74.7 & 82.3 & 73.1 & 83.8& 75.2 \\ \hline
Avg& 82.2 & 75.6 & 83.9 & 74.6 & 83.1 & 74.4 & 84.2& 75.4\\ \hline
\end{tabular}
\end{center}
\caption{An example of executing the experimental procedure applied to random search for Electra on RTE. The grid search accuracy is denoted using the {\bf\textcolor{blue}{blue bold font}}. An HPO run is highlighted in \colorbox{black!60}{dark grey if it overfits} and \colorbox{black!40}{medium grey if it overfits weakly}. \label{fig:walking_ex}}
\end{table}

\noindent \textbf{An Example of Executing the Procedure}. In Figure~\ref{fig:walking_ex}, we show an example of applying the procedure on random search for Electra on RTE. In round 0, the validation and test accuracies of all three repetitions are lower than grid search. That implies RS needs more time budget, therefore we increase the budget (marked as $\uparrow$res) for RS from 1GST to 4GST. After the increase, overfitting is detected in the 1st repetition of round 1 (validation accuracy = 84.5, test accuracy = 74.6). We thus reduce the search space (marked as $\downarrow$ space) from $\mathcal{S}_{full}$ to $\mathcal{S}_{-wr}$. In round 2, the 1st repetition still shows (weak) overfitting: RS has the same validation accuracy as grid search (84.1), a smaller test accuracy (76.1), and a smaller validation loss (RS's validation loss = 0.8233, grid search's validation loss = 0.9517). We thus continue reducing the search space to $\mathcal{S}_{min}$, and overfitting is detected again in the 1st repetition of round 3 (validation accuracy = 84.8, test accuracy = 75.3). After round 3, the search space cannot be further reduced, so we classify this case as 'HPO overfits task'.

We analyze the execution results in Table~\ref{file2:tab:electra_procedure_result} and \ref{file2:tab:roberta_procedure_result} jointly as follows.\\ 

\noindent \textbf{Effects of Reducing the Search Space}. From the two tables we can observe that reducing the search space can be effective for controlling overfitting. In WNLI (Electra), both algorithms outperform grid search after reducing the search space once. In WNLI (RoBERTa), ASHA outperforms grid search after reducing the search space twice. We can observe a similar trend in MRPC (Electra), SST (Electra), RTE (RoBERTa), and CoLA (RoBERTa). However, for these cases, overfitting still exists even after we reduce the search space twice, i.e., using the minimal search space. \\

\noindent \textbf{Effects of Increasing the Time Budget}. By observing cases of increased budget in Table~\ref{file2:tab:electra_procedure_result} and \ref{file2:tab:roberta_procedure_result}, we can see that this strategy is generally effective for improving the validation accuracy. After increasing the time budget, in STS-B (Electra) all HPO methods outperform grid search's validation and test accuracy; in SST (Electra-RS) and CoLA (RoBERTa) HPO outperforms grid search in only the validation accuracy. In RTE (Electra) and QNLI (Electra), however, this increase is not enough for bridging the gap with grid search, thus HPO remains behind. For RTE (Electra), SST (Electra), QNLI (Electra), and CoLA (RoBERTa), overfitting happens after increasing the time budget from 1GST to 4GST. After reducing the search space, we still observe overfitting in most cases. \\

\noindent \textbf{Comparisons between RS and ASHA}. By comparing the results between random search and ASHA in Table \ref{file2:tab:electra_procedure_result} and \ref{file2:tab:roberta_procedure_result}, we find that before increasing the budget, RS rarely outperforms ASHA in the validation accuracy; however, after the budget of both RS and ASHA increases to 4GST, the best validation accuracy of RS has consistently outperformed ASHA, i.e., in all of RTE (Electra), STS-B (Electra), SST (Electra), and QNLI (Electra). That is, the increase in the time budget has led to more significant (validation) increase in RS than ASHA. This result may be caused by two reasons. First, at 1GST, ASHA already samples a larger number of trials (Appendix~\ref{file2:sec:trial_numbers}), which may be sufficient to cover its search space; on the other hand, RS cannot sample enough trials, thus increasing the time budget is more helpful. Second, ASHA may make mistake by pruning a good trial that shows a bad performance at the beginning. 

\begin{table}[!ht]
\begin{center}
\begin{tabular}{lp{5.5cm}}
\hline { task } & { Execution Results } \\\hline \vspace{1.5pt} 
WNLI & \cellcolor{black!10}{All HPO succeed w/ 1GST, $\mathcal{S}_{-wr}$}\\
\multirow{2}{*}{RTE} &  \cellcolor{black!60}{RS overfits} \\
&  ASHA and BO+ASHA TBD \\\vspace{1.5pt}
MRPC & \cellcolor{black!60}{All HPO overfit} \\\vspace{1.5pt}
CoLA & \cellcolor{black!10}{All HPO succeed w/ 1GST, $\mathcal{S}_{full}$}\\ \vspace{1.5pt}
\multirow{1}{*}{STS-B} &  \cellcolor{black!10}{ All HPO succeed w/ 4GST, $\mathcal{S}_{full}$}\\\vspace{1.5pt}
SST & \cellcolor{black!60}{All HPO overfit} \\\vspace{1.5pt}
QNLI & All HPO TBD \\\vspace{1.5pt}
MNLI & \cellcolor{black!10}{All HPO succeed w/ 1GST, $\mathcal{S}_{full}$} \\
\hline
\end{tabular}
\end{center}

\begin{center}
\begin{tabular}{lp{6.0cm}}
\hline { task } & { Execution Results } \\ \hline
\multirow{2}{*}{WNLI} & \cellcolor{black!10}{ASHA succeeds\(^*\) w/ 1GST, $\mathcal{S}_{-wr-hdo}$}\\ \vspace{1.5pt}
& \cellcolor{black!60}{RS and BO+ASHA overfit} \\\vspace{1.5pt}
RTE & \cellcolor{black!60}{All HPO overfit} \\
\multirow{2}{*}{MRPC} & \cellcolor{black!10}{ASHA succeeds\(^*\) w/ 1GST, $\mathcal{S}_{-wr-hdo}$} \\\vspace{1.5pt}
& \cellcolor{black!60}{RS and BO+ASHA overfit} \\\vspace{1.5pt}
CoLA & \cellcolor{black!60}{All HPO overfit}\\
\multirow{2}{*}{STS-B} & \cellcolor{black!10}{RS succeeds w/ 1GST, $\mathcal{S}_{-wr-hdo}$} \\
& \cellcolor{black!10}{ASHA and BO+ASHA succeed w/ 1GST, $\mathcal{S}_{min}$} \\
\hline
\end{tabular}
\end{center}
\caption{Final results of executing the troubleshooting procedure on Electra (top) RoBERTa (bottom). \(^*\) means the risk of overfitting still exists based on the result of BO+ASHA\label{tab:exec_results_roberta}. }
\end{table}


\subsection{Summary of the Main Findings}

In Table~\ref{tab:exec_results_roberta}, we list the final execution results for each task in Electra and RoBERTa. Our main findings can be summarized as follows. After increasing the time budget and reducing the search space, HPO outperforms grid search in the following cases: (1) in 3 cases (i.e., CoLA (Electra), STS-B (Electra) and MNLI (Electra)), HPO outperforms grid search by using the full search space, where STS-B needs more budget; (2) in 4 cases (i.e., WNLI (Electra), WNLI (RoBERTa), MRPC (RoBERTa) and STS-B (RoBERTA)), HPO succeeds after reducing the search space; (3) in the other 7 cases, HPO cannot outperform grid search even after increasing the time budget and reducing the search space. This result shows that when searching in a continuous space surrounding the recommended grid configurations, it can be difficult for existing automated HPO methods (e.g., Random Search, ASHA, Bayesian optimization) to outperform grid search (with manually tuned grid configurations recommended by the language model) within a short amount of time; even if we can identify a configuration with good validation score, most likely the test score is still worse than grid search. \\

\noindent \textbf{The Total Running Time for the Procedure}. \noindent The execution for all experiments in Table~\ref{file2:tab:electra_procedure_result} and \ref{file2:tab:roberta_procedure_result} took 6.8$\times$4V100 GPU days. This is in contrast to the cost if we enumerate all 5 factors in Section~\ref{sec:env_factors}, which is 16×4V100 GPU days. \\

\noindent \textbf{A Caveat on Results in Table~\ref{tab:exec_results_roberta}}. For all study results in this paper (i.e., Table~\ref{tab:init_result_electra}, Table~\ref{file2:tab:electra_procedure_result} and Table~\ref{file2:tab:roberta_procedure_result}), we have repeated each HPO run three times. 
Therefore if a case succeed in Table~\ref{tab:exec_results_roberta}, it is because no overfitting is detected in the 3 repetitions, if we ran more repetitions, the risk of overfitting can increase. 
In addition, all results are evaluated under transformers version=3.4.0 and Ray version=1.2.0. If these versions change, results in Table~\ref{tab:exec_results_roberta} may change. \\

\noindent \textbf{An Analysis on the Relation between Overfitting and Train/Validation/Test split}. As overfitting indicates a negative correlation between the validation and test accuracy, one hypothesis is that overfitting is caused by the different distribution of the validation and test set. We thus compare HPO runs using the original GLUE spilt and a new split which uniformly partition the train/validation/test data. The results can be found in Appendix~\ref{sec:overfitting_analysis}. 

\section{Related Work}

\subsection{Automated Hyperparameter Optimization} 

Hyperparameter optimization methods for generic machine learning models have been studied for a decade~\cite{feurer2019hyperparameter,bergstra2011algorithms,bergstra2012random,swersky2013multi}. Prior to that, grid search was the most common tuning strategy~\cite{JMLR:Pedregosa2011}.
It discretizes the search space of the concerned hyperparameters and tries all the values in the grid. 
It can naturally take advantage of parallelism. However, 
The cost of grid search increases exponentially with hyperparameter dimensions.
A simple yet surprisingly effective alternative is to use random combinations of hyperparameter values, especially when the objective function has a low effective dimension, as shown in~\cite{bergstra2012random}.
Bayesian optimization (BO)~\cite{bergstra2011algorithms,snoek2012practical} fits a
probabilistic model to approximate the relationship between hyperparameter settings and their measured performance, uses this probabilistic model to make decisions about where next in the space to acquire
the function value, while integrating out uncertainty. Since the training of deep neural networks is very expensive, new HPO methods have been proposed to reduce the cost required. Early stopping methods \cite{karnin2013almost,li2017hyperband,Li2020} stop training with unpromising configurations at low fidelity (e.g., number of epochs) by comparing with other configurations trained at the same fidelity. 
Empirical study of these methods is mostly focused on the vision or reinforcement learning tasks, there has been few work focusing on NLP models. ASHA was evaluated on an LSTM model proposed in 2014~\cite{zaremba2014recurrent}. In \cite{wang2015efficient}, the authors empirically studied the impact of a multi-stage algorithm
for hyper-parameter tuning. In \cite{zhang2020reproducible}, a look-up table was created for hyperparameter optimization of neural machine translation systems. In BlendSearch~\cite{wang2021economic}, an economical blended search strategy was proposed to handle heterogeneous evaluation cost in general and demonstrates its effectiveness in fine-tuning a transformer model Turing-NLRv2.\footnote{\url{msturing.org}} Some existing work has addressed overfitting in HPO~\cite{levesque2018bayesian} or neural architecture search~\cite{zela2019understanding}. For HPO, cross validation can help alleviate the overfitting when tuning SVM~\cite{levesque2018bayesian}, which is rarely applied in deep learning due to high computational cost. For neural architecture search~\cite{zela2019understanding}, the solution also cannot be applied to our case due to the difference between the two problems. 




\subsection{Fine-tuning Pre-trained Language Models}

As fine-tuning pre-trained language models has become a common practice, existing works have studied how to improve the performance of the fine-tuning stage. Among them, many has focused on improving the robustness of fine-tuning. For example, ULMFit~\cite{howard2018universal} shows that an effective strategy for reducing the catastrophic forgetting in fine-tuning is to use the slanted triangular learning rate scheduler (i.e., using a non-zero number of warmup steps). Other strategies for controlling overfitting in fine-tuning include freezing a part of the layers to reduce the number of parameters, and gradually unfreezing the layers~\cite{peters2019tune},  adding regularization term to the objective function of fine-tuning~\cite{jiang2019smart}, multi-task learning~\cite{phang2018sentence}. Applying these techniques may reduce overfitting in our experiments; however, our goal is to compare grid search and HPO, if these techniques are helpful, they are helpful to both. To simplify the comparison, we thus focus on fine-tuning the original model. Meanwhile, the performance of fine-tuning can be significantly different with different choices of the random seeds~\cite{dodge2020fine}. To remove the variance from random seed, we have fixed all the random seeds to 42, although HPO can be used to search for a better random seed. \cite{zhang2021revisiting} identifies the instability of fine-tuning BERT model in few-sample cases of GLUE (i.e., RTE, MRPC, STS-B, and CoLA). Similar to our work, they also found that overfitting increases when searching for more trials. However, they have not compared grid search with HPO. There are also many discussions on how to control overfitting by tuning hyperparameters (in manual tuning), e.g., learning rate~\cite{smith2017bayesian}, batch size~\cite{smith2017don}, dropout rates~\cite{srivastava2014dropout,loshchilov2017decoupled,loshchilov2018fixing}, which may help with designing a search space for HPO that overfits less. 

\section{Conclusions, Discussions and Future Work}
\label{sec:discussion}

Our study suggests that for the problem of fine-tuning pre-trained language models, it is difficult for automated HPO methods to outperform manually tuned grid configurations with a limited time budget. However, it is possible to design a systematic procedure to troubleshoot the performance of HPO and improve the performance. We find that setting the search space appropriately per model and per task is crucial. Having that setting automated for different models and tasks is beneficial to achieve the goal of automated HPO for fine-tuning. For example, one may consider automatically mining the pattern from Table~\ref{file2:tab:reduce_sp_analysis_electra}\&\ref{file2:tab:reduce_sp_analysis_roberta} to identify the hyperparameters that likely cause overfitting.
Further, for the tasks remaining to be unsuitable for HPO, other means to reduce overfitting is required. One possibility is to use a different metric to optimize during HPO as a less overfitting proxy of the target metric on test data. 

Previous work has shown that random seed is crucial in the performance of fine-tuning~\cite{dodge2020fine}. Fine-tuning also benefits from ensembling or selecting a few of the best performing seeds~\cite{liu2019roberta}. 
It would be interesting to study HPO's performance by adding the random seed to the search space for future work.

In our study, the simple random search method stands strong against more advanced BO and early stopping methods. 
It suggests room for researching new HPO methods specialized for fine-tuning. A method that can robustly outperform random search with a small resource budget will be useful.

It is worth mentioning that although we find HPO sometimes underperforms grid search, the grid search configurations we study are the default ones recommended by the pre-trained language models for fine tuning, therefore they may be already extensively tuned. 
We may not conclude that HPO is not helpful when manual tuning has not been done. How to leverage HPO methods in that scenario is an open question. 

\clearpage
\bibliographystyle{acl_natbib}
\bibliography{acl2021}

\clearpage

\appendix
\section{Appendix}

\vspace{-5pt}
\subsection{HPO Checkpoint Settings}
\label{file2:sec:checkpoint}

In this paper, we report the validation and test accuracy of the \emph{best} checkpoint (in terms of validation accuracy) of the best trial instead of the \emph{last} checkpoint of the best trial. While the default setting in Ray Tune uses the last checkpoint, when fine-tuning pretrained language model without HPO, the best checkpoint is more widely used than the last checkpoint. To further study the difference between the two settings, we compare their validation and test accuracy of grid search using Electra on three tasks: WNLI, RTE and MRPC. The result shows that the validation and test accuracy of the \emph{best} checkpoint of the best trial are both higher than those of the \emph{last} checkpoint of the best trial. As a result, we propose and advocate to report the \emph{best} checkpoint of all the trials for HPO fine-tuning pretrained language models. The checkpoint frequencies in our experiment are set to 10 per epoch for larger tasks (SST, QNLI, and MNLI) and 5 per epoch for smaller tasks (WNLI, RTE, MRPC, CoLA and STS-B), with lower frequency in smaller tasks to reduce the performance drop caused by frequent I/Os within a short time. 

\subsection{Number of Trials Searched by HPO}
\label{file2:sec:trial_numbers}

In Table~\ref{file2:tab:trial_number}, we show the number of trials searched by each HPO algorithms in the initial comparative study ( Table~\ref{tab:init_result_electra}).

\begin{table}[h]
\begin{center}
\begin{tabular}{lccc}
\hline HPO & { RS } & ASHA & BO+ASHA \\ \hline
WNLI & 4 & 12 & 12 \\
RTE & 6 & 27 & 38 \\ 
MRPC & 5 & 36 & 36 \\
CoLA & 9 & 31 & 30 \\
STS-B & 4 & 31 & 33 \\
SST & 5 & 33 & 30 \\
QNLI & 4 & 26 & 24 \\
MNLI & 7 & 31 & 27 \\
\hline
\end{tabular}
\end{center}
\caption{Average numbers of trials searched by each HPO algorithm in the initial experiment on Electra. \label{file2:tab:trial_number} }
\end{table}

\vspace{-10pt}
\subsection{Choosing the Hyperparameter to Fix}
\label{file2:sec:reduce_sp_analysis}

The hyperparameters of the best trials in overfiting runs are shown in Table~\ref{file2:tab:reduce_sp_analysis_electra} and Table~\ref{file2:tab:reduce_sp_analysis_roberta}. We use colors to denote the comparison with the hyperparameter value in grid search: \colorbox{black!60}{dark grey} if the value is higher than grid search; \colorbox{black!10}{light grey} if the value is lower than grid search.

\subsection{Execution Results of Procedure}
\label{file2:exec_results_procedure}

In Table~\ref{file2:tab:electra_procedure_result} and Table~\ref{file2:tab:roberta_procedure_result}, we show the execution results of applying the experimental procedure to Electra and RoBERTa respectively. 

\subsection{An Analysis on Overfitting and Train/Validation/Test split}
\label{sec:overfitting_analysis}

In this paper, we have observed that HPO tends to overfit when the number of trials/time budget increases. In other words, the higher the validation score, the lower the test score. One hypothesis for the reason behind this phenomenon is that the validation set has a different distribution than the test set. Since GLUE is a collection of NLP datasets from different sources, it is unclear whether the validation and test set in all GLUE tasks share the same distribution.

\begin{table}[h]
\begin{center}
\begin{tabular}{cc|cc}
\hline
\multicolumn{2}{c|}{Origin} & \multicolumn{2}{c}{Resplit}\\ 
\hline validation  & test & validation & test \\ \hline
93.3& 93.3&91.9&91.8\\
93.2&93.2&91.7&91.6\\
93.2&93.1&91.6&91.1\\
93.1&93.4&91.6&91.5\\
\hline
\end{tabular}
\end{center}
\caption{Comparison of the orders of validation and test scores for the original split of GLUE and resplit. \label{tab:compare_origin_resplit} }
\vspace{-5pt}
\end{table}

\noindent To observe whether HPO still overfits under a uniformly random split, we have performed the following experiment: we merge the training and validation folds of QNLI in GLUE, randomly shuffle the merged data, and resplit it into train/validation/test with the proportion 8:1:1. We run random search, rank all trials based on the validation accuracy, and examine the Pearson correlation coefficient between the top-4 trials's validation and test accuracies (the trials are ranked by the validation accuracy), which are listed in Table~\ref{tab:compare_origin_resplit}. For the original GLUE dataset, we also save the best checkpoints of the top 4 trials and submit them to the GLUE website to get the test accuracies. The Pearson coefficient of the original dataset is \((r=-0.1414,p=0.858)\) while for resplit it is \((r=0.6602,p=0.339)\). Thus one potential explanation of the observed overfitting in this work is due to different distribution between validation and test data. 

\begin{table*}[!h]
\begin{center}
\begin{tabular}{lccccccc}
\hline HPO run & val acc & test acc & lr & wr & bs & hidd. do & att. do \\
\hline  
MRPC, grid & 92.3/89.2 & 91.1/87.9 & 1.0e-4 & 0.100 & 32 & 0.100 & 0.100 \\ 
MRPC, RS, rep 1 & 92.7/90.0 & 90.4/87.1 & {3.9e-5} & \cellcolor{black!10}{0.014} & 16 & \cellcolor{black!10}{0.050} & \cellcolor{black!10}{0.063} \\
MRPC, RS, rep 2 & 93.4/90.9 & 90.6/87.6 & {4.3e-5} & \cellcolor{black!10}{0.005} & 16 & \cellcolor{black!10}{0.044} & \cellcolor{black!10}{0.024} \\
MRPC, ASHA, rep 1 & 92.8/90.0 & 90.8/87.6 & {6.5e-5} & \cellcolor{black!10}{0.075} & 16 & \cellcolor{black!10}{0.038} & \cellcolor{black!10}{0.090}  \\ 
MRPC, ASHA, rep 2 & 93.4/90.9 & 90.5/87.4 & {3.1e-5} & \cellcolor{black!10}{0.030} & 16 & \cellcolor{black!10}{0.067} & \cellcolor{black!10}{0.097}  \\ 
MRPC, ASHA, rep 3 & 92.9/90.0 & 90.4/86.9 & {1.3e-4} & \cellcolor{black!10}{0.066} & 32 & \cellcolor{black!10}{0.097} & \cellcolor{black!10}{0.015}  \\ 
MRPC, Opt+ASHA, rep 1 & 93.0/90.4 & 90.7/87.5 & {6.4e-5} & \cellcolor{black!10}{0.084} & 16 & \cellcolor{black!60}{0.196} & \cellcolor{black!10}{0.002} \\
MRPC, Opt+ASHA, rep 2 & 93.3/90.7 & 90.4/86.9 & {8.0e-5} & \cellcolor{black!10}{0.010} & 32 & \cellcolor{black!10}{0.031} & \cellcolor{black!60}{0.108} \\
 \hline\hline
SST, grid & 95.1 & 95.7 & 3.0e-5 & 0.100 & 32 & 0.100 & 0.100 \\
SST, RS, rep 1 & 95.4 & 95.6 & {3.1e-5} & \cellcolor{black!10}{0.011} & 32 & \cellcolor{black!10}{0.006} & \cellcolor{black!10}{0.044} \\
 \hline\hline
STS-B, grid & 91.5/91.4 & 89.7/89.2 & 1.0e-4 & 0.100 & 32 & 0.100 & 0.100\\
STS-B, Opt+ASHA, rep 1 & 91.6/91.4 & 89.6/89.1 & {4.7e-5} & \cellcolor{black!10}{0.015} & 32 & \cellcolor{black!10}{0.028} & \cellcolor{black!10}{0.082}  \\
\hline

\end{tabular}
\end{center}
\caption{Comparison between the hyperparameter values of the best trial of grid search and the best trials (in validation accuracy) of all the 9 overfitting HPO runs (out of 72) in the initial comparative study using Electra (Table~\ref{tab:init_result_electra}). \colorbox{black!60}{dark grey} indicates the value is higher than grid search; \colorbox{black!10}{light grey} indicates the value is lower than grid search \label{file2:tab:reduce_sp_analysis_electra}}
\end{table*}

\begin{table*}[!h]
\begin{center}
\begin{tabular}{lcccccccc}
\hline HPO run & val acc & test acc & lr & wr & bs & hidd. do & att. do & wd \\
\hline  
WNLI,grid & 56.3 & 65.1 & - & 0.060 & - & {0.100} & 0.100 & 0.100 \\
WNLI,RS,rep 3 & 60.6 & 64.4 & {1.8e-5} & \cellcolor{black!60}{0.111} & 16 & \cellcolor{black!60}{0.128} & \cellcolor{black!60}{0.122} & \cellcolor{black!10}{0.078} \\
\hline\hline
CoLA,grid & 65.1 & 61.7 & 3.0e-5 & 0.060 & 16 & {0.100} & 0.100 & 0.100 \\
CoLA,ASHA, rep 1 & 65.5 & 59.5 & {2.7e-5} & \cellcolor{black!10}{0.020} & 32 & \cellcolor{black!10}{0.090} & \cellcolor{black!60}{0.197} & \cellcolor{black!60}{0.180} \\
CoLA,Opt+ASHA,rep 1 & 65.4 & 59.4 & {2.3e-5} & \cellcolor{black!60}{0.067} & 32 & \cellcolor{black!10}{0.063} & \cellcolor{black!60}{0.117} & \cellcolor{black!60}{0.293} \\
\hline\hline
RTE,grid & 79.8 & 73.9 & 3.0e-5 & 0.060 & 16 & {0.100} & 0.100 & 0.100 \\
RTE,RS,rep 1 & 80.5 & 73.6 & {2.8e-5} & \cellcolor{black!60}{0.085} & 16 & \cellcolor{black!10}{0.025} & \cellcolor{black!60}{0.173} & \cellcolor{black!60}{0.142} \\
RTE,ASHA,rep 3 & 80.5 & 73.2 & {2.4e-5} & \cellcolor{black!10}{0.022} & 16 & \cellcolor{black!10}{0.053} & \cellcolor{black!60}{0.137} & \cellcolor{black!10}{0.016} \\
RTE,Opt+ASHA,rep 2 & 81.9 & 73.5 & {2.7e-5} & \cellcolor{black!10}{0.024} & 32 & \cellcolor{black!10}{0.083} & \cellcolor{black!60}{0.190} & \cellcolor{black!10}{0.094} \\ 
\hline\hline
MRPC,grid & 93.1/90.4 & 90.5/87.1 & 2.0e-5 & 0.060 & 16 & {0.100} & 0.100 & 0.100 \\
MRPC,RS,rep 2 & 93.2/90.7 & 89.6/86.1 & {2.4e-5} & \cellcolor{black!60}{0.094} & 64 & \cellcolor{black!10}{0.019} & \cellcolor{black!60}{0.138} & \cellcolor{black!60}{0.299} \\
MRPC,RS,rep 3 & 93.2/90.4 & 90.3/86.7 & {1.4e-5} & \cellcolor{black!10}{0.003} & 16 & \cellcolor{black!10}{0.011} & \cellcolor{black!10}{0.062} & \cellcolor{black!60}{0.176} \\
MRPC,ASHA,rep 3 & 93.3/90.7 & 90.3/86.8 & {2.7e-5} & \cellcolor{black!10}{0.008} & 16 & \cellcolor{black!60}{0.140} & \cellcolor{black!60}{0.130} & \cellcolor{black!60}{0.255} \\
MRPC,Opt+ASHA,rep 3 & 93.5/91.2 & 89.6/86.2 & {2.7e-5} & \cellcolor{black!10}{0.036} & 16 & \cellcolor{black!10}{0.094} & \cellcolor{black!60}{0.153} & \cellcolor{black!60}{0.291} \\
\hline\hline
STS-B,grid & 91.2/90.8 & 89.3/88.4 & 2.0e-5 & 0.060 & 16 & {0.100} & 0.100 & 0.100 \\
STS-B,ASHA,rep 1 & 91.3/91.0 & 89.0/88.2 & {2.0e-5} & \cellcolor{black!10}{0.042} & 16 & \cellcolor{black!10}{0.004} & \cellcolor{black!10}{0.061} & \cellcolor{black!60}{0.247} \\
STS-B,ASHA,rep 2 & 91.4/91.1 & 89.0/88.2 & {2.1e-4} & \cellcolor{black!60}{0.061} & 16 & \cellcolor{black!10}{0.056} & \cellcolor{black!10}{0.008} & \cellcolor{black!60}{0.226} \\
STS-B,Opt+ASHA,rep 1 & 91.3/90.9 & 89.1/88.2 & {2.7e-5} & \cellcolor{black!10}{0.052} & 16 & \cellcolor{black!10}{0.096} & \cellcolor{black!10}{0.070} & \cellcolor{black!60}{0.224} \\
\hline
\end{tabular}
\end{center}
\caption{Comparison between the hyperparameter values of the best trial of grid search and the best trials (in validation accuracy) of all the 11 overfitting HPO runs (out of 45) in the initial comparative study using RoBERTa (Table~\ref{tab:init_result_electra}). \colorbox{black!60}{dark grey} indicates the value is higher than grid search; \colorbox{black!10}{light grey} indicates the value is lower than grid search  \label{file2:tab:reduce_sp_analysis_roberta}}
\end{table*}

\begin{table*}[!h]
\begin{center}
\begin{tabular}{l|p{0.45cm}p{0.45cm}|p{0.45cm}p{0.45cm}|p{0.45cm}p{0.45cm}|p{0.45cm}p{0.45cm}||p{0.45cm}p{0.45cm}|p{0.45cm}p{0.45cm}|p{0.45cm}p{0.45cm}|p{0.45cm}p{0.45cm}}
\hline &  \multicolumn{8}{c|}{Random Search} & \multicolumn{8}{c}{ASHA}  \\\cline{2-17}   
&  \multicolumn{2}{c|}{round 0} & \multicolumn{2}{c|}{round 1}  &  \multicolumn{2}{c|}{round 2} & \multicolumn{2}{c||}{round 3}  &  \multicolumn{2}{c|}{round 0} & \multicolumn{2}{c|}{round 1}  &  \multicolumn{2}{c|}{round 2} & \multicolumn{2}{c}{round 3} \\
& val & test & val & test & val & test & val & test & val & test & val & test & val & test & val & test \\ \hline
\multirow{4}{*}{WNLI} & {\bf\textcolor{blue}{56.3}} & {\bf\textcolor{blue}{65.1}} & \multicolumn{2}{c|}{$\downarrow$ space} &  &  &  &  & \multicolumn{2}{c|}{} & \multicolumn{2}{c|}{$\downarrow$ space} &  &  &  & \\ \cline{2-17}
& \cellcolor{black!60}{57.7} & \cellcolor{black!60}{62.3} & 57.7 & 65.8 &  &  &  &   & \cellcolor{black!60}{57.7} & \cellcolor{black!60}{63.0} & 59.2 & 65.8 &  &  &  & \\
& 56.3 & 65.8 & 57.7 & 65.1 &  &  &  &  & \cellcolor{black!60}{57.7} & \cellcolor{black!60}{59.6} & 57.7 & 65.1 &  &  &  & \\
& 56.3 & 65.1 & 57.7 & 65.1 &  &  & & & 56.3 & 65.1 & 57.7 & 65.8 &  &  & & \\ \cline{2-17}
& 56.8 & 64.4 & \cellcolor{black!10}{57.7} & \cellcolor{black!10}{65.3} &  &  & & & 57.2 & 62.6 & \cellcolor{black!10}{58.2} & \cellcolor{black!10}{65.6} &  &  & & \\ \hline

\multirow{4}{*}{RTE} &{\bf\textcolor{blue}{84.1}} & {\bf\textcolor{blue}{76.8}} & \multicolumn{2}{c|}{$\uparrow$ res} &  \multicolumn{2}{c|}{$\downarrow$ space} &  \multicolumn{2}{c|}{$\downarrow$ space} &  \multicolumn{2}{c|}{} &  \multicolumn{2}{c|}{$\uparrow$ res}  &  &  & \\ \cline{2-17}
& 81.9 & 76.1 & \cellcolor{black!60}{84.5} & \cellcolor{black!60}{74.6} & \cellcolor{black!40}{84.1} & \cellcolor{black!40}{76.1} & \cellcolor{black!60}{84.8} &  \cellcolor{black!60}{75.3} & 81.9& 76.2 & 83.4 & 75.3 &  &  &  & \\
& 81.6 & 75.1 & 83.8 & 74.5 & 83.0 &  74.0 & 84.1 & 75.7 & 75.5 &72.1 & 81.9 & 73.9 & & &  & \\
& 83.0 & 75.7 & 83.4 & 74.7 & 82.3 & 73.1 &83.8 & 75.2& 83.4 & 74.1& 83.8 & 74.4 &  &  & & \\ \cline{2-17}
& 82.2 & 75.6 & 83.9 & 74.6 & 83.1 & 74.4 &84.2 & 75.4& 80.3& 74.1 & 83.0 & 74.5 &  &  & & \\ \hline

\multirow{4}{*}{MRPC} & {\bf\textcolor{blue}{89.2}} & {\bf\textcolor{blue}{87.9}} & \multicolumn{2}{c|}{$\downarrow$ space} & \multicolumn{2}{c|}{$\downarrow$ space} &  & & \multicolumn{2}{c|}{$\downarrow$ space} & \multicolumn{2}{c|}{$\downarrow$ space} &  &  &  & \\ \cline{2-17}
& \cellcolor{black!60}{90.9} & \cellcolor{black!60}{87.6} & \cellcolor{black!60}{90.7} & \cellcolor{black!60}{86.3} & \cellcolor{black!60}{90.4} & \cellcolor{black!60}{86.5} &  &   & \cellcolor{black!60}{90.9} & \cellcolor{black!60}{87.4} &  \cellcolor{black!60}{90.0} & \cellcolor{black!60}{87.2} & \cellcolor{black!60}{90.2} & \cellcolor{black!60}{87.6} &  &  \\
& \cellcolor{black!60}{90.0} & \cellcolor{black!60}{87.1} & 90.2 & 87.2 & 90.7 & 86.5 &  &  & \cellcolor{black!60}{90.0} & \cellcolor{black!60}{86.9} & 90.4 & 87.8 & 90.9 & 88.3 &  & \\
& 90.2 & 87.8 & 90.7 & 86.9 & 90.7 & 87.8 & & & \cellcolor{black!60}{90.0} & \cellcolor{black!60}{87.6} & 89.5 & 86.0 & 90.7 & 87.6 & & \\ \cline{2-17}
& 90.4 & 87.5 & 90.5 & 86.8 & 90.6 & 87.4 & & & \cellcolor{black!60}{90.3} & \cellcolor{black!60}{87.3} & 90.4 & 87.0 & 90.6 & 87.8 & & \\ \hline

\multirow{4}{*}{STS-B} & {\bf\textcolor{blue}{91.4}} & {\bf\textcolor{blue}{89.2}} & \multicolumn{2}{c|}{$\uparrow$ res} & & &  & & \multicolumn{2}{c|}{} & \multicolumn{2}{c|}{$\uparrow$ res} &  &  &  & \\ \cline{2-17}
& 90.8 & 89.1 & 91.5 & 89.4 &  &  &  &   & 91.3 & 89.2 & 91.5 & 89.8 &  &  &  &  \\
& 89.6 & 85.9 &91.4 & 89.6 &  &  &  &  & 91.5& 89.7 & 91.4 & 89.2 &  &  &  & \\
& 90.1 & 87.7 & 91.5 & 89.9 & &  & & & 91.0 & 88.3 & 91.4 & 89.2 &  &  & & \\ \cline{2-17}
& 90.2 & 87.6 & \cellcolor{black!10}{91.4} & \cellcolor{black!10}{89.6} &  &  & & & 91.3 & 89.1 & \cellcolor{black!10}{91.4} & \cellcolor{black!10}{89.4} &  &  & & \\ \hline

\multirow{4}{*}{SST} & {\bf\textcolor{blue}{95.1}} & {\bf\textcolor{blue}{95.7}} & \multicolumn{2}{c|}{$\downarrow$ space} & \multicolumn{2}{c|}{$\uparrow$ res} &  \multicolumn{2}{c|}{$\downarrow$ space} & \multicolumn{2}{c|}{} & \multicolumn{2}{c|}{$\uparrow$ res} & \multicolumn{2}{c|}{$\downarrow$ space}  & \multicolumn{2}{c}{$\downarrow$ space}   \\ \cline{2-17}
& \cellcolor{black!60}{95.4} & \cellcolor{black!60}{95.6} & 93.2 & 93.8 & \cellcolor{black!60}{96.0} & \cellcolor{black!60}{94.7} & \cellcolor{black!60}{95.6} & \cellcolor{black!60}{95.2}  & 95.4 & 95.8 & \cellcolor{black!60}{95.5} & \cellcolor{black!60}{95.3} & \cellcolor{black!60}{95.5} & \cellcolor{black!60}{95.2} &  \cellcolor{black!60}{95.2} &  \cellcolor{black!60}{94.9} \\
& 94.3 & 95.1 &94.7 & 95.0 & 95.3 & 95.7 & 95.1 & 95.7 & 94.4 & 94.1 & 95.1 & 94.7 & 94.8 & 94.3 & 94.2 & 93.6\\
& 94.5 & 94.6 & 95.8 & 95.7 & 95.5 & 95.8 & 95.0 & 94.5& 95.0 & 94.9 & \cellcolor{black!60}{95.4} & \cellcolor{black!60}{95.4} & 94.5 & 93.5 & 94.8& 94.5\\ \cline{2-17}
& 94.7 & 95.1 & 94.6 & 94.8 & 95.6 & 95.4 & 95.2 & 95.1 & 94.9 & 94.9 & 95.3 & 95.1 & 94.9 & 94.3 & 94.7 & 94.3 \\ \hline

\multirow{4}{*}{QNLI} & {\bf\textcolor{blue}{93.5}} & {\bf\textcolor{blue}{93.5}} & \multicolumn{2}{c|}{$\uparrow$ res} & & &   & & \multicolumn{2}{c|}{} & \multicolumn{2}{c|}{$\uparrow$ res} & & &  & \\ \cline{2-17}
& 93.0 & 92.9 &  93.2 & 93.4 & &  &  &   & 92.5 & 92.4 & 93.4 & 93.2 & & &  &  \\
& 93.1 & 93.6 & 93.3 & 93.3 &  &  &  &  & 93.4 & 93.0 & 93.2 & 93.1 &  &  &  & \\
& 92.9 & 92.5 & 93.3 & 93.1 & &  & & & 93.4 & 93.4 & 93.2 & 93.0 &  &  & & \\ \cline{2-17}
& 93.0 & 93.0 & 93.3 & 93.3 &  &  & & & 93.1 & 92.9 & 93.3 & 93.1 &  &  & & \\ \hline

\end{tabular}
\end{center}
\caption{The execution results of applying the procedure on Electra. Each task's grid search accuracy is denoted using the {\bf\textcolor{blue}{blue bold font}}. An HPO run is highlighted in \colorbox{black!60}{dark grey if it overfits} and \colorbox{black!40}{medium grey if it overfits weakly}. The average of 3 repetitions is highlighted in \colorbox{black!10}{light grey if it outperforms grid search's validation and test accuracy}. For STS-B we only report the Spearman correlation, for MRPC we only report the accuracy. \label{file2:tab:electra_procedure_result}}
\end{table*}

\begin{table*}[!h]
\begin{center}
\begin{tabular}{l|p{0.45cm}p{0.45cm}|p{0.45cm}p{0.45cm}|p{0.45cm}p{0.45cm}|p{0.45cm}p{0.45cm}||p{0.45cm}p{0.45cm}|p{0.45cm}p{0.45cm}|p{0.45cm}p{0.45cm}|p{0.45cm}p{0.45cm}}
\hline  &  \multicolumn{8}{c|}{Random Search} & \multicolumn{8}{c}{ASHA}  \\\cline{2-17}
&  \multicolumn{2}{c|}{round 0} & \multicolumn{2}{c|}{round 1}  &  \multicolumn{2}{c|}{round 2} & \multicolumn{2}{c||}{round 3}  &  \multicolumn{2}{c|}{round 0} & \multicolumn{2}{c|}{round 1}  &  \multicolumn{2}{c|}{round 2} & \multicolumn{2}{c}{round 3} \\
& val & test & val & test & val & test & val & test & val & test & val & test & val & test & val & test \\ \hline
\multirow{4}{*}{WNLI} & {\bf\textcolor{blue}{56.3}} & {\bf\textcolor{blue}{65.1}} & \multicolumn{2}{c|}{$\downarrow$ space} & \multicolumn{2}{c|}{$\downarrow$ space} &    &  & \multicolumn{2}{c|}{} & \multicolumn{2}{c|}{$\downarrow$ space} &  \multicolumn{2}{c|}{$\downarrow$ space} &  & \\ \cline{2-17}
& \cellcolor{black!60}{60.6} & \cellcolor{black!60}{64.4} & \cellcolor{black!60}{62.0} & \cellcolor{black!60}{64.4} & \cellcolor{black!60}{57.7} & \cellcolor{black!60}{62.3} &  &   & \cellcolor{black!40}{59.2} & \cellcolor{black!40}{65.1} & \cellcolor{black!40}{59.2} & \cellcolor{black!40}{65.1} & 57.7 & 65.8 &  & \\
& 56.3 & 65.1 & 56.3 & 65.1 & 56.3 & 65.1 &  &  & 56.3 & 65.1& 56.3 & 65.1 & 56.3 & 65.1 &  & \\
& 56.3 & 65.1 & 56.3 & 65.1 & 56.3 & 65.1 & & & 56.3 & 65.1 & 56.3 & 65.1 & 56.3 & 65.1 & & \\ \cline{2-17}
& 57.8 & 64.9 & 58.2 & 64.9 & 56.8 & 64.2 & & & 57.3  & 65.1 & 57.3 & 65.1 & \cellcolor{black!10}{56.8} & \cellcolor{black!10}{65.3} & & \\ \hline

\multirow{4}{*}{RTE} & {\bf\textcolor{blue}{79.8}} & {\bf\textcolor{blue}{73.9}} & \multicolumn{2}{c|}{$\downarrow$ space} &  \multicolumn{2}{c|}{$\downarrow$ space} & & &  \multicolumn{2}{c|}{} &   \multicolumn{2}{c|}{$\downarrow$ space} & \multicolumn{2}{c|}{$\downarrow$ space} &  & \\ \cline{2-17}
& 81.2 & 73.9 & \cellcolor{black!60}{80.1} & \cellcolor{black!60}{72.8} & \cellcolor{black!60}{81.6} & \cellcolor{black!60}{72.2} &  &   &\cellcolor{black!60}{80.5} & \cellcolor{black!60}{73.2} &  \cellcolor{black!60}{80.5} & \cellcolor{black!60}{73.3} & \cellcolor{black!40}{79.8} & \cellcolor{black!40}{72.5} &  & \\
& \cellcolor{black!60}{80.5}& \cellcolor{black!60}{73.6} & \cellcolor{black!60}{81.2} & \cellcolor{black!60}{72.9} & 75.5 &  72.1&  & & 80.2 & 74.9& \cellcolor{black!60}{82.0} & \cellcolor{black!60}{72.9} & 79.1 & 73.4 &  & \\
& 79.4 & 73.1 & \cellcolor{black!40}{79.8} & \cellcolor{black!40}{73.6} & \cellcolor{black!40}{79.8} & \cellcolor{black!40}{72.6} & & &80.5  &74.1 & \cellcolor{black!60}{80.5} & \cellcolor{black!60}{73.5} & 79.8 & 73.7 & & \\ \cline{2-17}
& 80.4 & 73.5 & 80.4 & 73.1 & 78.9 & 72.3 & & &80.8& 74.1 & \cellcolor{black!60}{80.5} & \cellcolor{black!60}{73.3} & 79.5 & 73.2 & & \\ \hline

\multirow{4}{*}{MRPC} & {\bf\textcolor{blue}{90.4}} & {\bf\textcolor{blue}{87.1}} & \multicolumn{2}{c|}{$\downarrow$ space} & \multicolumn{2}{c|}{$\downarrow$ space} &  & & & & \multicolumn{2}{c|}{$\downarrow$ space} & &&  & \\ \cline{2-17}
& \cellcolor{black!60}{90.7} & \cellcolor{black!60}{86.1} & \cellcolor{black!60}{90.7} & \cellcolor{black!60}{86.9} & \cellcolor{black!60}{91.2} & \cellcolor{black!60}{86.7} &  &   & \cellcolor{black!60}{90.7} & \cellcolor{black!60}{86.8} &  91.4 & 87.7 &  &  &  &  \\
& \cellcolor{black!40}{90.4} & \cellcolor{black!40}{86.7} & 90.4 & 88.0 & 90.2 & 87.6 &  &  & 90.4 & 87.4 & 90.4 & 87.2 &  &  &  & \\
& 90.9 & 87.2 & 91.2 & 87.2 & \cellcolor{black!40}{90.4} & \cellcolor{black!40}{87.0} & & & 91.4 & 87.6 & 90.4& 87.6  &  &  & & \\ \cline{2-17}
& 90.7 & 86.7 & 90.8 & 87.4 & 90.6 & 87.1 &  & & 90.8 & 87.3 & \cellcolor{black!10}{90.8} & \cellcolor{black!10}{87.5} & &  &  & \\ \hline

\multirow{4}{*}{CoLA} & {\bf\textcolor{blue}{65.1}} & {\bf\textcolor{blue}{61.7}} & \multicolumn{2}{c|}{$\uparrow$ res} &  \multicolumn{2}{c|}{$\downarrow$ space} & \multicolumn{2}{c|}{$\downarrow$ space} &  \multicolumn{2}{c|}{} &   \multicolumn{2}{c|}{$\downarrow$ space} & \multicolumn{2}{c|}{$\downarrow$ space} & \\ \cline{2-17}
& 64.3 & 60.1 & \cellcolor{black!60}{66.0} & \cellcolor{black!60}{59.3} & \cellcolor{black!60}{65.8} & \cellcolor{black!60}{59.2} &  \cellcolor{black!60}{65.3}& \cellcolor{black!60}{60.2}  &\cellcolor{black!60}{65.5} & \cellcolor{black!60}{59.5} & 65.0 & 60.9 & \cellcolor{black!60}{65.9} & \cellcolor{black!60}{58.2} &  & \\
& 64.6& 60.5 & 65.0 & 60.5 & 65.0 &  61.7& 65.4 & 62.5 & 63.6& 58.8& 62.9& 58.4 &  63.9&  58.9&  & \\
& 63.5 & 56.8 & 64.4 & 60.3 & 65.2 & 60.7 & 64.6 & 58.5&64.6  &60.0 & \cellcolor{black!60}{64.9} & \cellcolor{black!60}{62.0} & 64.4 & 59.0 & & \\ \cline{2-17}
& 64.1 & 59.1 & 65.1 & 60.0 & 65.3 & 60.5 & 65.1 & 60.4&64.5 & 59.4 & 64.3 & 60.4& 64.7 & 58.7 & & \\ \hline

\multirow{4}{*}{STS-B} &{\bf\textcolor{blue}{90.8}} & {\bf\textcolor{blue}{88.4}} & \multicolumn{2}{c|}{$\downarrow$ space}  & & &  & & \multicolumn{2}{c|}{} & \multicolumn{2}{c|}{$\downarrow$ space} & \multicolumn{2}{c|}{$\downarrow$ space}  &  \\ \cline{2-17}
& \cellcolor{black!60}{90.8} & \cellcolor{black!60}{88.3} & 91.0 & 88.9 &  &  &  &   & \cellcolor{black!60}{91.1} & \cellcolor{black!60}{88.2} & \cellcolor{black!60}{90.9} & \cellcolor{black!60}{88.3} & 90.8 & 88.6  &  &  \\
& 90.8 & 88.9 &90.8 & 88.6 &  &  &  &  & \cellcolor{black!60}{91.0} & \cellcolor{black!60}{88.2} & 90.8 & 88.5 & 91.0 & 88.5 &  & \\
& 91.2 & 88.7 & 90.9 & 88.9 &  &  & & & 90.7 & 88.5 & 90.9 & 88.4 & 90.9 & 88.7 & & \\ \cline{2-17}
& 90.9 & 88.6 & \cellcolor{black!10}{90.9} & \cellcolor{black!10}{88.8} &  &  & & &90.9& 88.3&  90.8 & 88.4 & \cellcolor{black!10}{90.9} & \cellcolor{black!10}{88.6} & & \\ \hline

\end{tabular}
\end{center}
\caption{The execution results of applying the procedure on RoBERTa. Each task's grid search accuracy is denoted using the {\bf\textcolor{blue}{blue bold font}}. An HPO run is highlighted in \colorbox{black!60}{dark grey if it overfits} and \colorbox{black!40}{medium grey if it overfits weakly}. The average of 3 repetitions is highlighted in \colorbox{black!10}{light grey if it outperforms grid search's validation and test accuracy}. For STS-B we only report the Spearman correlation, for MRPC we only report the accuracy. \label{file2:tab:roberta_procedure_result}}
\end{table*}


\end{document}